\documentclass{article}

\usepackage[preprint]{neurips_2025}

\usepackage[utf8]{inputenc} 
\usepackage[T1]{fontenc}    
\usepackage{hyperref}       
\usepackage{url}            
\usepackage{booktabs}       
\usepackage{amsfonts}       
\usepackage{nicefrac}       
\usepackage{microtype}      
\usepackage{xcolor}         

\usepackage{graphicx}
\usepackage{amsmath}
\usepackage{CJKutf8}

\title{PHRASED: Phrase Dictionary Biasing for Speech Translation}

\author{%
  Peidong Wang \\
  Microsoft\\
  USA \\
  \texttt{peidongwang@microsoft.com} \\
  \And
  Jian Xue \\
  Microsoft\\
  USA \\
  \texttt{jian.xue@microsoft.com} \\
  \And
  Rui Zhao \\
  Microsoft\\
  USA \\
  \texttt{ruzhao@microsoft.com} \\
  \And
  Junkun Chen \\
  Microsoft\\
  USA \\
  \texttt{junkunchen@microsoft.com} \\
  \And
  Aswin Shanmugam Subramanian \\
  Microsoft\\
  USA \\
  \texttt{aswins@microsoft.com} \\
  \And
  Jinyu Li \\
  Microsoft\\
  USA \\
  \texttt{jinyli@microsoft.com} \\
}

\begin{document}

\maketitle

\begin{abstract}
Phrases are essential to understand the core concepts in conversations. However, due to their rare occurrence in training data, correct translation of phrases is challenging in speech translation tasks. In this paper, we propose a phrase dictionary biasing method to leverage pairs of phrases mapping from the source language to the target language. We apply the phrase dictionary biasing method to two types of widely adopted models, a transducer-based streaming speech translation model and a multimodal large language model. Experimental results show that the phrase dictionary biasing method outperforms phrase list biasing by 21\% relatively for the streaming speech translation model. In addition, phrase dictionary biasing enables multimodal large language models to use external phrase information, achieving 85\% relative improvement in phrase recall.
\end{abstract}

\section{Introduction}
\label{sec:intro}
Phrases contain important information in conversations \citep{niehues2017exploiting,li2018named,hassan2018achieving,post2019exploration}. However, they are constantly emerging and evolving and are typically not seen in the training set, posing challenges to modern speech processing systems, especially speech translation (ST) models that takes speech as input and directly outputs translated texts in another language.

There are generally three types of ST models, cascaded, end-to-end (E2E), and multimodal large language model (LLM), each of which has different methods for phrase translation.
Cascaded ST systems consist of an automatic speech recognition (ASR) model followed by a text-based machine translation (MT) model \citep{Ney1999ST, Matusov2005ST, Post2013ST}. The modular design makes it easy to use the phrase translation methods for MT. However, cascaded ST systems have multiple issues. First, errors in ASR may propagate to MT. Second, cascaded ST systems cannot fully leverage speech information (e.g., prosody) for translation. Finally, cascaded systems typically have long inference latency.
E2E systems have already established superiority over cascaded systems \citep{li2022recent, prabhavalkar2023end}. E2E ST, which directly maps speech features in one language to texts in another language, has been widely adopted \citep{vila2018end,sperber2020speech}, especially in applications requiring streaming capability \citep{xue2022large,wang2022lamassu,xue2023weakly}. The translation of phrases in E2E ST systems is harder than that in cascaded systems. Gaido \emph{et al.} analyzed such named entity errors \citep{gaido2022we}. They proposed joint speech translation and named entity recognition methods \citep{gaido2023joint}, and named entity detection and injection methods \citep{gaido2023named}. 
To improve the performance of Conformer-transducer (CT)-based ST models \citep{xue2022large}, Zhao \emph{et al.} proposed a connectionist temporal classification (CTC) guided modality matching method CTC-GMM \citep{zhao2024ctc}. To leverage MT training data, it adds a CTC compression module to the encoder to align speech and text modalities. 
Note that since the CTC compression module is jointly trained with the other parts of the E2E ST model and is used the same way as a standard CT encoder, CTC-GMM may still be considered an E2E ST model.
Recently, multimodal LLMs such as GPT-4o \citep{hurst2024gpt}, Gemini \citep{team2023gemini}, and Phi-4-multimodal \citep{abouelenin2025phi}, are able to perform ST tasks. 

In this paper, we focus on the phrase translation problem of transducer-based streaming ST models and multimodal LLMs. Conventionally, there are generally two ways to improve phrase translation \citep{gaido2022we}, model fine-tuning, and biasing. Model fine-tuning uses utterances containing phrases to update all or part of the model parameters so that the phrases are exposed to the model. The advantage of model fine-tuning is that it does not require external information during inference. However, it is difficult for model fine-tuning to deal with the phrases that are newly emerged or fast evolving. In addition, after fine-tuning, the model stores the phrase information in model weights, resulting in a potential risk of data leakage.
For ASR tasks, model biasing methods typically use a phrase list provided by the user \citep{zhao2019shallow,wang2024contextual}. During inference, a biasing module increases the probability of the occurrence of those phrases in the ASR output. For ST tasks, it is natural for users to provide pairs of source-language and target-language phrases to specify the desired translation behavior. To exploit such source-target phrase pairs information, we propose \textbf{phrase} \textbf{d}ictionary biasing (PHRASED). PHRASED can be used for phrase list selection or joint biasing. Experimental results of PHRASED show significant improvement over phrase list biasing (PLB) for both transducer-based streaming ST models and multimodal LLMs.


The remainder of this paper is organized as follows. In Section \ref{sec:related}, we list related works. We illustrate the proposed PHRASED method in Section \ref{sec:sys}. In Sections \ref{sec:exp} and \ref{sec:res}, we describe the experimental setup and evaluation results. We conclude in Section \ref{sec:conc}.

\section{Related Work}
\label{sec:related}

\textbf{Phrase dictionary biasing for MT.} It is straightforward for MT models to use phrase dictionaries for biasing, since both their input and output are texts. Phrase dictionary biasing for MT has been extensively explored. Some representative methods include placeholder-based methods \citep{wang2017sogou,post2019exploration,li2019neural}, special tokens \citep{li2018named,modrzejewski2020incorporating}, code-switching \citep{song2019code}, entity embeddings \citep{sennrich2016linguistic,niehues2017exploiting,ugawa2018neural}, and end-to-end entity-aware models \citep{xie2022end}. However, these methods cannot be easily used for ST models, especially the E2E streaming ST models and multimodal LLMs in this study. The reason is that different from MT, the input modality of ST is speech instead of text, making it difficult to define the source-language phrase representation.





\noindent \textbf{Retrieval augmented generation.} Retrieval augmented generation (RAG) is a popular method to use external information to control the output of LLMs \citep{zhao2024retrieval}. Du \emph{et al.} proposed to decompose the ST task into sequential steps of speech recognition and translation \citep{du2024cot}. Hu \emph{et al.} proposed to perform chain-of-thought prompting for ST \citep{hu2025chain}. These methods focused on improving the overall translation quality of the whole utterance and not phrase translation. Another field of RAG related to PHRASED is generative error correction (GEC) \citep{ma2023can,yang2023generative,ghosh2024failing}, where an LLM is used to correct the output of a first-pass model. However, most prior arts in this field focused on ASR tasks.

\section{Method Description}
\label{sec:sys}

\subsection{Phrase List Biasing for ST}
\label{ssec:phrase_list_biasing}
Let us denote the acoustic observations of an ST model as $\mathbf{x}=(\mathbf{x}_1, ...,\mathbf{x}_T)$, where $T$ is the total frame length. An ST model generates the posterior probabilities $P(\mathbf{y}|\mathbf{x})$ for a set of subword units $\mathbf{y}=(\mathbf{y}_1, ...,\mathbf{y}_L)$, where $L$ is the length of the output.

\begin{equation}
    \mathbf{y}^{ST} = \mathop{\arg\max}\limits_{\mathbf{y}} \log P(\mathbf{y}|\mathbf{x})
\end{equation}

We use $\mathbf{O} = [\mathbf{O}_1,...,\mathbf{O}_N]$ to denote the external phrase list in the target language, where $N$ denotes the total number of phrases. The shallow fusion-based phrase list biasing method interpolates the score of the ST model with the external phrase list $\mathbf{O}$ during decoding.

\begin{equation}
    \mathbf{y}^{*} = \mathop{\arg\max}\limits_{\mathbf{y}} \log P(\mathbf{y}|\mathbf{x}) + \lambda \log P_\mathbf{O}(\mathbf{y})
\end{equation}
where $\mathbf{y}^*$ denotes the output after applying the biasing method, $\lambda \geq 0$ is a tunable hyperparameter controlling how much the contextual language model influences the final output, and $P_\mathbf{O}(\mathbf{y})$ is the corresponding score from the phrase list biasing module.



\subsection{Phrase Dictionary Biasing for ST}
\label{ssec:PHRASED}
For ST tasks, when specifying the biasing phrases in the target language, the corresponding phrases in the source language are typically also provided. We denote such source-language phrase list as $\mathbf{I} = [\mathbf{I}_1, ..., \mathbf{I}_N]$. $\mathbf{I}$ can be various types of representation, including audio, text, and intermediate embeddings. 
With phrase dictionary $\{\mathbf{I}:\mathbf{O}\}$, PHRASED can be denoted as below.

\begin{equation}
    \mathbf{y}^{*} = \mathop{\arg\max}\limits_{\mathbf{y}} \log P(\mathbf{y}|\mathbf{x}) + \lambda \log P_{\{\mathbf{I}:\mathbf{O}\}}(\mathbf{y})
\end{equation}
where $P_{\{\mathbf{I}:\mathbf{O}\}}(\mathbf{y})$ is the score from the phrase biasing module using the $\{\mathbf{I}:\mathbf{O}\}$ dictionary.

There are multiple ways to leverage the $\{\mathbf{I}:\mathbf{O}\}$ dictionary. We discuss two methods in this paper, phrase selection, and joint biasing.

\subsubsection{Phrase Selection}
\label{sssec:phrase_selection}
Phrase selection is to select the target-language phrases where the corresponding source-language phrases appear in the intermediate representation $\mathbf{z} = (\mathbf{z}_1, ..., \mathbf{z}_K)$ of an utterance, where $K$ is the length of the intermediate representation. In this paper, $\mathbf{z}$ denotes ASR texts. We refer to the source-language phrases in $\mathbf{I}$ that appear in $\mathbf{z}$ as $\mathbf{I}^m = (\mathbf{I}^m_1, ..., \mathbf{I}^m_U)$, where $U$ is the number of matched phrases for the specific utterance. The corresponding target-language phrases can be denoted as $\mathbf{O}^m = (\mathbf{O}^m_1, ..., \mathbf{O}^m_U)$. $\mathrm{PHRASED}_{phrase\_selection}$ can be expressed as follows.

\begin{equation}
    \mathbf{y}^{*} = \mathop{\arg\max}\limits_{\mathbf{y}} \log P(\mathbf{y}|\mathbf{x}) + \lambda \log P_{\mathbf{O}}(\mathbf{y}) + \mu \log P_{\mathbf{O}^m}(\mathbf{y|z})
    \label{eq:phrase_selection}
\end{equation}
where $\mu \geq 0$ is the hyperparameter controlling the influence of the selected target-language phrases $\mathbf{O}^m$, and $P_{\mathbf{O}^m}(\mathbf{y|z})$ is the corresponding score of PHRASED using phrase selection.
Note that the $\{\mathbf{I}:\mathbf{O}\}$ dictionary information is implicitly used during the phrase matching and selection process, where only the $\mathbf{O}^m$ corresponding to $\mathbf{I}^m$ are used for biasing. It is important to point that the biasing weights $\lambda$ and $\mu$ can be zeros. $\lambda = 0$ should be used when the recall of the ASR model is $100\%$. When $\mu = 0$, the biasing method defaults back to the simple phrase list biasing method.

\subsubsection{Joint Biasing}
\label{sssec:llm_based_joint_biasing}
Phrase selection cannot exploit the information in the phrase dictionary since $\mathbf{I}$ is not explicitly used. We denote the method that explicitly uses both source-language and target-language phrases as $\mathrm{PHRASED}_{joint\_biasing}$, which can be expressed by the equation below.
\begin{equation}
    \mathbf{y}^{*} = \mathop{\arg\max}\limits_{\mathbf{y}} \log P(\mathbf{y}|\mathbf{x}) + \lambda \log P_{\{\mathbf{I}:\mathbf{O}\}}(\mathbf{y}) + \mu \log P_{\{\mathbf{I}^m: \mathbf{O}^m\}}(\mathbf{y|z})
    \label{eq:joint_biasing}    
\end{equation}

Compared with phrase selection, joint biasing has more modeling power, but its applicability depends on the specific model structure. We discuss the usage of PHRASED for streaming ST and multimodal LLMs in the following subsections.

\subsection{PHRASED for Streaming ST}
\label{ssec:PHRASED_streaming}
For streaming ST, we adopt CTC-GMM, a transducer-based model containing a CTC compression module in its encoder. The CTC compression module is pre-trained using a large amount of multilingual ASR data, therefore can naturally serve as the intermediate representation $\mathbf{z}$. We adopt the phrase selection variant of PHRASED as shown in equation (\ref{eq:phrase_selection}) for streaming ST.

\begin{figure*}[th]
    \centering
    \includegraphics[width=0.96\linewidth]{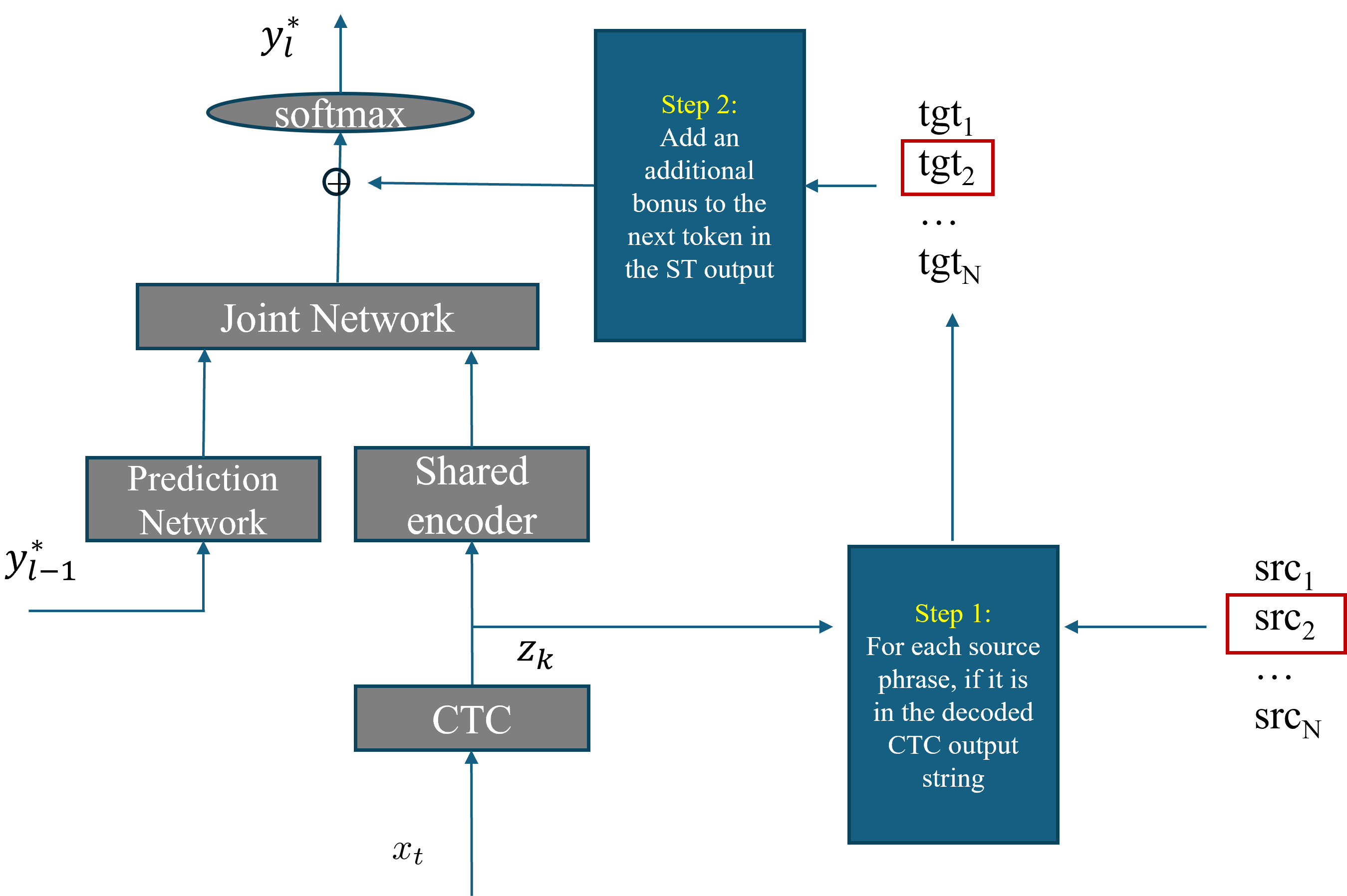}
    \caption {Illustration of $\mathrm{PHRASED}_{phrase\_selection}$ for CTC-GMM. At Step 1, we find the source-language phrases that occur in the intermediate representation. At Step 2, we add bonuses to the ST model output for the selected target-language phrases.}
    \label{ref:fig_ctc}
\end{figure*}

Figure \ref{ref:fig_ctc} explains the $\mathrm{PHRASED}_{joint\_biasing}$ method for CTC-GMM. At each decoding step, we obtain the ASR result from the CTC output by converting the byte-level byte-pair encoding (BBPE) tokens to text. 
Then for the phrases in the source-language phrase list, we compare them with the decoding result. If a source-language phrase is found in the decoding result, we add a bonus to the corresponding target-language phrase at the ST model output. Note that due to the output reordering nature of ST, once a target-language phrase is selected, it will impact all the following decoding steps.

To add the PHRASED bonus to the ST model output, we keep track of the word piece indices of the selected target-language phrases that are partially matched with $\mathbf{y}$. For all potential next ST output tokens at a decoding step, if they match any of the word pieces corresponding to the current indices of the selected target-language phrases, we add an PHRASED bonus to that output token dimension. 
If later we find that some phrases are only partially matched with $\mathbf{y}$, we need to remove the bonuses that have already been added to the output to clean up the output scores. To calculate the bonus value we need to remove, we multiply the index of the longest partially matched target-language phrase and the per-token PHRASED bonus.

\subsection{PHRASED for Multimodal LLM}
\label{ssec:PHRASED_llm}

\begin{figure}[th]
    \centering
  \includegraphics[width=0.96\linewidth]{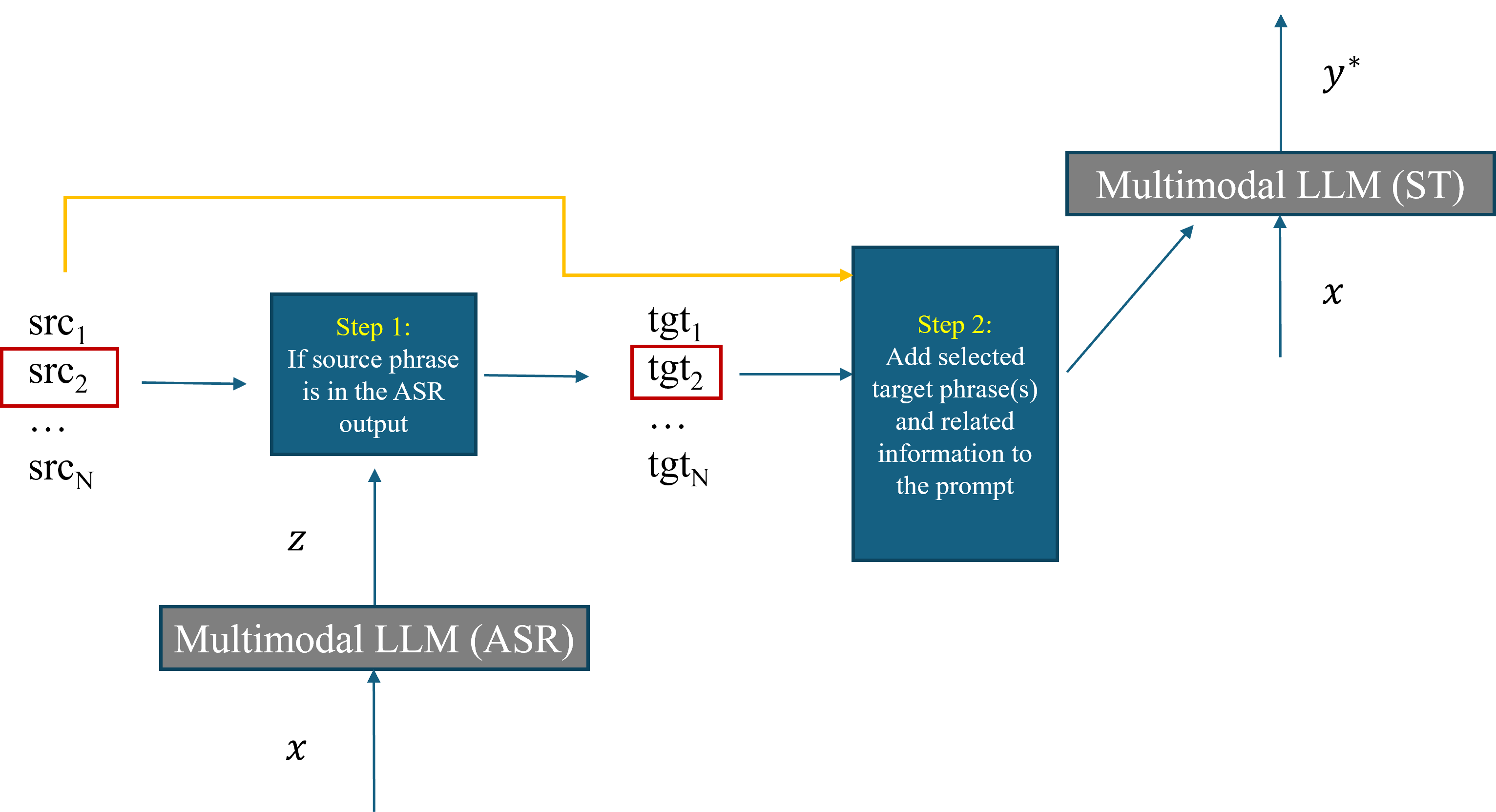}
  \caption {Illustration of $\mathrm{PHRASED}_{joint\_biasing}$ for multimodal LLMs. At Step 1, we find the source-language phrases that occur in the intermediate representation $\mathbf{z}$. At Step 2, we add the selected target-language phrases and related intermediate representation information to the prompt.}
    \label{ref:fig_llm}
\end{figure}

A multimodal LLM can support various speech functions in a single model. We first let the multimodal LLM convert audio to text using its ASR functionality. Then we explore two ways to improve its phrase translation quality: phrase selection and joint biasing.
Figure \ref{ref:fig_llm} illustrates the joint biasing method for multimodal LLMs.
Note that current multimodal LLMs cannot easily use phrase lists in prompts since the number of phrases is very large in real-world applications. Therefore, we set $\lambda = 0$ to equations (\ref{eq:phrase_selection}) and (\ref{eq:joint_biasing}).
\begin{equation}
    \mathbf{y}^{*} = \mathop{\arg\max}\limits_{\mathbf{y}} \log P(\mathbf{y}|\mathbf{x}) + \mu \log P_{\mathbf{O}^m}(\mathbf{y|z})
    \label{eq:phrase_selection_simplified}
\end{equation}
\begin{equation}
    \mathbf{y}^{*} = \mathop{\arg\max}\limits_{\mathbf{y}} \log P(\mathbf{y}|\mathbf{x}) + \mu \log P_{\{\mathbf{I}^m: \mathbf{O}^m\}}(\mathbf{y|z})
    \label{eq:joint_biasing_simplified}
\end{equation}

For equations (\ref{eq:phrase_selection_simplified}) and (\ref{eq:joint_biasing_simplified}), the additional prompts we used are ``The output should contain $\mathbf{O}^m$.'' and ``The $\mathbf{I}^m$ in the audio clip should be translated to $\mathbf{O}^m$.'', respectively.




\section{Experimental Setup}
\label{sec:exp}

\subsection{Baseline Models}
\label{ssec:exp_models}
We use three types of ST models in this study, CT, CTC-GMM, and multimodal LLM.

\subsubsection{CT}
\label{sssec:exp_model_ct}
The CT model contains three components, an encoder, a prediction network, and a joint network. The total number of parameters in the CT model is 400M.
In the encoder, the number of Conformer layers is 42. For each Conformer layer, the attention dimension is 512 and the number of attention heads is 8. The dimension of the linear units is 2048. We do not use batch normalization and adopt swish as the activation function.
For the prediction network, we use 4 layers of long short-term memory (LSTM) cells. For each layer, the size of the hidden state is 1536 and the input size is 320. We use a dropout rate of 0.1 during training.
The joint network combines the output of the encoder and the prediction network. The joint dimension is 512 and the output dimension is 4332. 

The CT model performs ST in a streaming manner. The system latency of the CT model is 1 second and can leverage 18 seconds of history information.

The CT model is trained on 351K hours of anonymized weakly supervised in-house training data generated by translating multilingual ASR transcriptions to English using a text-based MT model.
During training, we set peak learning rate to 0.0003, the number of warm-up steps to 1M, and the total step to 28M. The optimizer is AdamW with betas [0.9, 0.98]. 
DeepSpeed is used with gradient clipping at 0.01, FP16 enabled, and zero optimization set to stage 2. 
We use 32 V100 graphical processing units (GPUs) to train the model.

\subsubsection{CTC-GMM}
\label{sssec:exp_model_ctc}
The CTC-GMM model adds a CTC compression module on top of the encoder of the CT model. 
The number of Conformer layers in the speech encoder is 26. The attention dimension, number of attention heads, and dimension of linear units are the same as in the CT model.
The CTC compression module has 16 Conformer layers. The embedding dimension is 512 and the output dimension is 30002. It uses BBPE to tokenize the multilingual ASR transcription. The chunk size of the CTC compression module is 8 and the number of left chunks is 18.
The configurations of the prediction network and the joint network are the same as those of the CT model.

The CTC-GMM model is also a streaming ST model. The system latency is also 1 second and can access 18 seconds of audio history information.

To train the CTC-GMM model, in addition to the ST training data, we use the text-based MT data of Japanese, Italian, Korean, French, Spanish, Portuguese, Chinese, and Germany. Such text-based training data exposes the CTC-GMM model with more phrases. During training, we set the peak learning rate to 0.0004, with 1M warm-up steps and 54M total steps. To train the model, we use a combination of Transducer loss and CTC loss, with loss weights of 1.0 and 0.1, respectively. Different from CT, the CTC-GMM model does not apply DeepSpeed and only uses the standard PyTorch distributed data parallel (DDP) package.


\subsubsection{Multimodal LLM}
\label{sssec:exp_model_phi}
For the multimodal LLM, we adopt Phi-4-multimodal, an open-source 5.6B model supporting text, vision, and audio modalities. The prompt to perform ASR is ``Transcribe the audio clip into text.'' and that for ST is ``Translate the audio clip to English.''. Phi-4-multimodal requires A100 GPUs for decoding.

\subsection{Variants of PHRASED}
\label{ssec:exp_variants}

\begin{table}[htb]
  \centering
  \begin{tabular}{lcc}
    \hline
    \textbf{methods} & $\mathrm{PHRASED}_{phrase\_selection}$ & $\mathrm{PHRASED}_{joint\_biasing}$ \\
    \hline
    CTC-GMM & \checkmark &  \\
    Phi-4-multimodal & \checkmark & \checkmark \\
    \hline
  \end{tabular}
  \caption{Two variants of PHRASED for CTC-GMM and Phi-4-multimodal. The methods investigated in this paper are marked as $\checkmark$.}
  \label{tab:variants}
\end{table}

As mentioned in Section \ref{ssec:PHRASED}, there are two variants of PHRASED, phrase selection, and joint biasing. In this paper, we apply these two variants to CTC-GMM and Phi-4-multimodal, as shown in Table \ref{tab:variants}. The joint biasing variant for CTC-GMM may involve a specific encoder module to leverage the source-language phrases. We leave it for future work.

\subsection{Evaluation Method}
\label{ssec:exp_data}
We calculate the bilingual evaluation understudy (BLEU) scores and phrase recalls of the models on the Chinese to English subset of RealSI \citep{cheng2024towards}. 
BLEU scores are used mainly for translation quality evaluation, but can also be an indirect indicator of phrase recall. The reason is that correct phrase translation may also improve n-grams.
The phrase recall of ST is calculated slightly different from that of ASR. If a phrase is not a filler word and the ASR is verbatim, there is a clear correspondence between the occurrence of a phrase in the audio input and that in the reference text. 
However, for ST evaluation, there are multiple valid translation results and some may use pronouns to refer to the phrase.
Therefore, when calculating the phrase recall in this paper, we only count the target-language phrase once, regardless of the number of its occurrence. Note that with this phrase recall calculation method, the recall numbers are typically lower in absolute value compared with counting with duplications. However, it should not impact the comparisons among different methods in this paper.


The test set covers 10 domains: technology, health, education, finance, law, environment, entertainment, science, and art. For each utterance, it provides human-annotated phrases in both Chinese and English.
We segmented the test audio based on the human-annotated start and end time.

To create a more realistic phrase list, we added irrelevant phrases randomly sampled from OntoNote5 \citep{pradhan2013towards}, which contains phrases from various domains. The total number of phrases in the phrase list is 3K. 
All the results in this paper are with the 3K phrase list, unless stated otherwise.

\section{Evaluation Results}
\label{sec:res}

\subsection{PHRASED for Streaming ST}
\label{ssec:res_phrased_streaming}



Table \ref{tab:ctc} shows the BLEU and phrase recall comparisons of different methods for streaming ST models. Compared the CT baseline, the CTC-GMM baseline improves the BLEU score by 1.8 and phrase recall by 33\% relatively. This indicates that the CTC-GMM model leverages the phrases in the text-only MT training data. Phrase list biasing improves the phrase recalls of both CT and CTC-GMM by 50\% relatively, showing the effectiveness of model biasing methods.

\begin{table}[htb]
  \centering
  \begin{tabular}{lcc}
    \hline
    \textbf{methods} & \textbf{BLEU} & \textbf{recall} \\
    \hline
    CT baseline & 16.5 & 21.62\% \\
    PLB & \underline{\textbf{19.2}} & \underline{\textbf{32.43\%}} \\
     \hline
    CTC-GMM baseline & 18.3 & 28.83\% \\
    PLB & 19.9 & 43.24\% \\
    $\mathrm{PHRASED}_{phrase\_selection}$ & \underline{\textbf{20.0}} & \underline{\textbf{52.25\%}} \\  
    PLB w/ large bonus & 4.6 & 49.55\% \\
    \hline
  \end{tabular}
  \caption{BLEU and phrase recall comparisons of different methods for the two streaming ST models, CT and CTC-GMM. PLB is short for phrase list biasing. The best results in each column are bolded and underlined.}
  \label{tab:ctc}
\end{table}


With the CTC compression module, CTC-GMM is able to use the proposed PHRASED method. Compared with phrase list biasing, PHRASED improves the recall by 21\% relatively. 
Since PHRASED introduces additional 3.0 biasing bonus to the original 1.0 phrase list biasing bonus, we add an ablation study using 4.0 as the bonus for phrase list biasing. 
The BLEU score drops to 4.6 and the recall is smaller than PHRASED. This shows the importance of phrase selection.

\subsection{Impact of Phrase Lists}
\label{ssec:phrase_list_type}

\begin{table}[htb]
  \centering
  \begin{tabular}{llcc}
    \hline
    \textbf{phrase list} & \textbf{methods} & \textbf{BLEU} & \textbf{recall} \\
    \hline
    oracle dataset-wise & PLB & 19.8 & 58.56\% \\
                 & $\mathrm{PHRASED}_{phrase\_selection}$ & 19.5 & 64.86\% (11\%) \\  
    \hline
    3K phrases & PLB & 19.9 & 43.24\% \\
         & $\mathrm{PHRASED}_{phrase\_selection}$ & {{20.0}} & {{52.25\%}} (21\%) \\  
    \hline
  \end{tabular}
  \caption{BLEU and phrase recall comparisons of PLB and PHRASED for CTC-GMM using different phrase lists. Oracle dataset-wise denotes the phrase list only containing phrases that occur in the test dataset, whereas 3K phrases has phrases irrelevant from the test set. In real-world applications, it is hard to obtain the oracle dataset-wise phrase list. The percentage numbers in the braces show the relative improvements over the corresponding PLB methods.}
  \label{tab:ctc_phrase_list}
\end{table}

Table \ref{tab:ctc_phrase_list} shows the impact of phrase lists on the comparisons between phrase list biasing and PHRASED. With the more realistic 3K phrase list containing irrelevant phrases, the improvement of PHRASED becomes larger, showing the importance of phrase dictionary biasing in real-world applications.

\subsection{Sample Output Sentences}
\label{ssec:sample_outputs}

\begin{table}[htb]
  \centering
  \begin{tabular}{l p{0.7\linewidth}@{}}
    \hline
    \textbf{methods} & \textbf{output} \\
    \hline
    reference & Of course, the \colorbox{yellow!50}{marketing and distribution} can be handed over to another company. This is also possible. \\
    CTC-GMM baseline &  Of course, it is also possible that this movie will be given to another company, which is also possible. \\
    PLB &  Of course, it is also possible that this movie will be given to another company, which is also possible. \\
    $\mathrm{PHRASED}_{phrase\_selection}$ & Of course, it is also possible that this movie will give the \colorbox{yellow!50}{marketing and distribution} to another company, which is also possible. \\  

    \hline
    reference & Therefore, I think screening \colorbox{yellow!50}{YouTube} videos. \\
    CTC-GMM baseline & In this situation, I think it's very important to understand. \\
    PLB & In this situation, I think it's very important to understand that. \\
    $\mathrm{PHRASED}_{phrase\_selection}$ & In this situation, the screening of \colorbox{yellow!50}{YouTube} videos. \\  
    
    \hline
  \end{tabular}
  \caption{Sample output sentences of different biasing methods for CTC-GMM. The phrase dictionary of the two sample output sentences are \{\begin{CJK*}{UTF8}{gbsn}宣传和发行\end{CJK*}: \colorbox{yellow!50}{marketing and distribution}\} and \{YouTube: \colorbox{yellow!50}{YouTube}\}, respectively.}
  \label{tab:ctc_samples}
\end{table}

Table \ref{tab:ctc_samples} contains sample output sentences of different methods.
PHRASED is able to translate the phrases both CTC-GMM and phrase list biasing cannot handle.
More importantly, the phrases in the table alter the meaning of the output sentences, leading to potentially severe consequences. 
In the first sample output, ``giving the movie to another company'' and ``giving the marketing and distribution of the movie to another company'' have very different meanings. For the second sample output sentence, the missing ``YouTube'' also greatly impacts the understanding of the sentence. These sample output sentences demonstrate the benefit of using PHRASED.

\subsection{PHRASED for Multimodal LLM}
\label{ssec:res_phrased_llm}

Table \ref{tab:phi4} shows the BLEU and phrase recall comparisons of different methods for Phi-4-multimodal. Compared with the CT and CTC-GMM baselines in Table \ref{tab:ctc}, the Phi-4-multimodal baseline performs better in both BLEU and phrase recall. This is expected as Phi-4-multimodal is larger in size and is trained on a large amount of text data. When applying conventional phrase list biasing methods to Phi-4-multimodal using prompts, we observe that the model cannot perform ST and only outputs ``I'm sorry, but the provided text appears to be a random string of words and numbers without any clear meaning or context. Could you please provide more information or clarify what you are looking for?''. In fact, we observe similar behaviors in other multimodal LLMs such as GPT-4o. This may be because most LLMs are not trained to take as large as 3K phrases in the prompt. 

\begin{table}[htb]
  \centering
  \begin{tabular}{lcc}
    \hline
    \textbf{methods} & \textbf{BLEU} & \textbf{recall} \\
    \hline
    Phi-4-multimodal baseline & 21.1 & 36.04\% \\
    PLB & 1.1 & 8.11\% \\  
    $\mathrm{PHRASED}_{phrase\_selection}$ & 23.8 & 54.95\% \\  
    $\mathrm{PHRASED}_{joint\_biasing}$ & \underline{\textbf{24.0}} & \underline{\textbf{66.67\%}} \\
    \hline
  \end{tabular}
  \caption{BLEU and phrase recall comparisons of different methods for Phi-4-multimodal.}
  \label{tab:phi4}
\end{table}

When we apply $\mathrm{PHRASED}_{phrase\_selection}$, the LLM is able to use the external phrase information. The BLEU score improves by 2.7 and the recall improves by 52\% relatively. 
With $\mathrm{PHRASED}_{joint\_biasing}$, the improvement in BLEU is 2.9 and that in recall is 85\% relatively. This suggests that the LLM learns the mapping from the source-language phrases to the corresponding target-language phrases and such phrase dictionary information is indeed useful for phrase translation.

\section{Concluding Remarks}
\label{sec:conc}
We have proposed PHRASED for ST. By using not only the target-language phrases but also the corresponding source-language phrases, we obtain 21\% relative improvement over phrase list biasing for the streaming ST model CTC-GMM. In addition, PHRASED enables a multimodal LLM to perform phrase biasing, achieving 2.9 improvement in BLEU score and 85\% relative improvement in phrase recall.
These results demonstrate that using phrase dictionaries in ST is beneficial, paving the road to bridge the language barrier.



\section{Limitations}
\label{sec:limitations}
As part of future work, we would like to work on the following limitations of PHRASED. 

\begin{itemize}
    \item We only explored ASR texts as the source-language phrase representation. There may be better representations such as audio embeddings.
    \item We only used text prompt to add the biasing information to the multimodal LLM. There are other ways that may be more effective, such as biasing on the LLM output.
    \item We only evaluated the results on the Chinese-English translation direction due to the limited human-annotated test sets for phrase translation. We may collect more data for our own evaluation and make it available to the research community.
    \item For CTC-GMM, we only applied the phrase selection variant of PHRASED. We may design specific methods to leverage the source-language phrases for it.
    \item We used an offline Phi-4-multimodal model. Reducing the latency of LLMs is important especially for ST tasks. We may design better streaming multimodal LLMs and investigate PHRASED for them.
    \item Current PHRASED for multimodal LLMs require two decoding steps. To reduce the inference cost, we may design single-pass decoding methods to apply PHRASED for multimodal LLMs.
\end{itemize}

\section{Societal Impacts}
\label{sec:societal}

\subsection{Potential Positive Societal Impacts}
\label{ssec:societal_pos}
The work presented in this paper has the following potential positive societal impacts.

\begin{itemize}
    \item Enhanced cross-cultural understanding and reduced miscommunication: The proposed method can improve ST performance. This facilitates communication and understanding between people speaking different languages and of different cultural backgrounds. In diplomacy, international business, and personal interactions, this can foster greater empathy, trust, and more successful collaborations.
    \item Increased access to specialized information and essential services: The proposed method significantly improves phrase recall. Fields like healthcare, law, and emergency response often use specific, complex terminology and set phrases. Better phrase translation recall would make ST systems more reliable in these critical contexts.
\end{itemize}



\subsection{Potential Negative Societal Impacts}
\label{ssec:socital_neg}
While improving phrase translation recall in ST systems has significant benefits, there are also potential negative societal impacts to consider.

\begin{itemize}
    \item The main concerns about model biasing or model customization in general may be the privacy and security risks. As mentioned in the paper, different from fine-tuning-based methods, the proposed method does not store user-specific information in the model. Although the method requires phrase dictionary inputs from users, the biasing data is stored locally and thus may not pose privacy or security risks. 
    \item As ST systems become highly proficient at translating common phrases and idioms, users may develop a strong sense of trust in their accuracy. However, even with improved recall, these systems can still struggle with highly novel phrases, sarcasm, irony, culturally specific allusions, or deliberately ambiguous language. To mitigate this issue, the ST systems may also provide their level of confidence in translating specific phrases. Another way is to incorporate deeper contextual understanding, speaker intent recognition, and the ability to detect and appropriately handle nuances like sarcasm or irony. We may also conduct user education or human oversight in critical applications.
\end{itemize}

\bibliographystyle{acl_natbib}
\bibliography{custom.bib}

\begin{thebibliography}{38}
\providecommand{\natexlab}[1]{#1}

\bibitem[{Abouelenin et~al.(2025)Abouelenin, Ashfaq, Atkinson, Awadalla, Bach, Bao, Benhaim, Cai, Chaudhary, Chen et~al.}]{abouelenin2025phi}
Abdelrahman Abouelenin, Atabak Ashfaq, Adam Atkinson, Hany Awadalla, Nguyen Bach, Jianmin Bao, Alon Benhaim, Martin Cai, Vishrav Chaudhary, Congcong Chen, et~al. 2025.
\newblock Phi-4-mini technical report: Compact yet powerful multimodal language models via mixture-of-loras.
\newblock \emph{arXiv preprint arXiv:2503.01743}.

\bibitem[{Cheng et~al.(2024)Cheng, Huang, Ko, Li, Peng, Xu, and Zhang}]{cheng2024towards}
Shanbo Cheng, Zhichao Huang, Tom Ko, Hang Li, Ningxin Peng, Lu~Xu, and Qini Zhang. 2024.
\newblock \href {https://arxiv.org/abs/2407.21646} {Towards achieving human parity on end-to-end simultaneous speech translation via llm agent}.
\newblock \emph{arXiv preprint arXiv:2407.21646}.

\bibitem[{Du et~al.(2024)Du, Ma, Yang, Deng, Chen, Yang, Xiang, Liu, and Qin}]{du2024cot}
Yexing Du, Ziyang Ma, Yifan Yang, Keqi Deng, Xie Chen, Bo~Yang, Yang Xiang, Ming Liu, and Bing Qin. 2024.
\newblock Cot-st: Enhancing llm-based speech translation with multimodal chain-of-thought.
\newblock \emph{arXiv preprint arXiv:2409.19510}.

\bibitem[{Gaido et~al.(2022)Gaido, Negri, and Turchi}]{gaido2022we}
Marco Gaido, Matteo Negri, and Marco Turchi. 2022.
\newblock Who are we talking about? handling person names in speech translation.
\newblock In \emph{Proceedings of the 19th International Conference on Spoken Language Translation (IWSLT 2022)}, pages 62--73.

\bibitem[{Gaido et~al.(2023{\natexlab{a}})Gaido, Papi, Negri, and Turchi}]{gaido2023joint}
Marco Gaido, Sara Papi, Matteo Negri, and Marco Turchi. 2023{\natexlab{a}}.
\newblock Joint speech translation and named entity recognition.
\newblock In \emph{Proc. Interspeech 2023}, pages 47--51.

\bibitem[{Gaido et~al.(2023{\natexlab{b}})Gaido, Tang, Kulikov, Huang, Gong, and Inaguma}]{gaido2023named}
Marco Gaido, Yun Tang, Ilia Kulikov, Rongqing Huang, Hongyu Gong, and Hirofumi Inaguma. 2023{\natexlab{b}}.
\newblock Named entity detection and injection for direct speech translation.
\newblock In \emph{ICASSP 2023-2023 IEEE International Conference on Acoustics, Speech and Signal Processing (ICASSP)}, pages 1--5. IEEE.

\bibitem[{Ghosh et~al.(2024)Ghosh, Rasooli, Levit, Wang, Xue, Manocha, and Li}]{ghosh2024failing}
Sreyan Ghosh, Mohammad~Sadegh Rasooli, Michael Levit, Peidong Wang, Jian Xue, Dinesh Manocha, and Jinyu Li. 2024.
\newblock Failing forward: Improving generative error correction for asr with synthetic data and retrieval augmentation.
\newblock \emph{arXiv preprint arXiv:2410.13198}.

\bibitem[{Hassan et~al.(2018)Hassan, Aue, Chen, Chowdhary, Clark, Federmann, Huang, Junczys-Dowmunt, Lewis, Li et~al.}]{hassan2018achieving}
Hany Hassan, Anthony Aue, Chang Chen, Vishal Chowdhary, Jonathan Clark, Christian Federmann, Xuedong Huang, Marcin Junczys-Dowmunt, William Lewis, Mu~Li, et~al. 2018.
\newblock Achieving human parity on automatic chinese to english news translation.
\newblock \emph{arXiv preprint arXiv:1803.05567}.

\bibitem[{Hu et~al.(2025)Hu, Chen, Yang, {\.Z}elasko, Hrinchuk, Lavrukhin, Balam, and Ginsburg}]{hu2025chain}
Ke~Hu, Zhehuai Chen, Chao-Han~Huck Yang, Piotr {\.Z}elasko, Oleksii Hrinchuk, Vitaly Lavrukhin, Jagadeesh Balam, and Boris Ginsburg. 2025.
\newblock Chain-of-thought prompting for speech translation.
\newblock In \emph{ICASSP 2025-2025 IEEE International Conference on Acoustics, Speech and Signal Processing (ICASSP)}, pages 1--5. IEEE.

\bibitem[{Hurst et~al.(2024)Hurst, Lerer, Goucher, Perelman, Ramesh, Clark, Ostrow, Welihinda, Hayes, Radford et~al.}]{hurst2024gpt}
Aaron Hurst, Adam Lerer, Adam~P Goucher, Adam Perelman, Aditya Ramesh, Aidan Clark, AJ~Ostrow, Akila Welihinda, Alan Hayes, Alec Radford, et~al. 2024.
\newblock Gpt-4o system card.
\newblock \emph{arXiv preprint arXiv:2410.21276}.

\bibitem[{Li(2022)}]{li2022recent}
Jinyu Li. 2022.
\newblock Recent advances in end-to-end automatic speech recognition.
\newblock \emph{APSIPA Transactions on Signal and Information Processing}, 11(1).

\bibitem[{Li et~al.(2019)Li, Yan, Zhang, and Zong}]{li2019neural}
Xiaoqing Li, Jinghui Yan, Jiajun Zhang, and Chengqing Zong. 2019.
\newblock Neural name translation improves neural machine translation.
\newblock In \emph{Machine Translation: 14th China Workshop, CWMT 2018, Wuyishan, China, October 25-26, 2018, Proceedings 14}, pages 93--100. Springer.

\bibitem[{Li et~al.(2018)Li, Wang, Aw, Chng, and Li}]{li2018named}
Zhongwei Li, Xuancong Wang, AiTi Aw, Eng~Siong Chng, and Haizhou Li. 2018.
\newblock Named-entity tagging and domain adaptation for better customized translation.
\newblock In \emph{Proceedings of the seventh named entities workshop}, pages 41--46.

\bibitem[{Ma et~al.(2023)Ma, Qian, Manakul, Gales, and Knill}]{ma2023can}
Rao Ma, Mengjie Qian, Potsawee Manakul, Mark Gales, and Kate Knill. 2023.
\newblock Can generative large language models perform asr error correction?
\newblock \emph{arXiv preprint arXiv:2307.04172}.

\bibitem[{Matusov et~al.(2005)Matusov, Kanthak, and Ney}]{Matusov2005ST}
Evgeny Matusov, Stephan Kanthak, and Hermann Ney. 2005.
\newblock On the integration of speech recognition and statistical machine translation.
\newblock In \emph{Interspeech}, pages 3177--3180.

\bibitem[{Modrzejewski et~al.(2020)Modrzejewski, Exel, Buschbeck, Ha, and Waibel}]{modrzejewski2020incorporating}
Maciej Modrzejewski, Miriam Exel, Bianka Buschbeck, Thanh-Le Ha, and Alex Waibel. 2020.
\newblock Incorporating external annotation to improve named entity translation in nmt.
\newblock In \emph{Proceedings of the 22nd annual conference of the european association for machine translation}, pages 45--51.

\bibitem[{Ney(1999)}]{Ney1999ST}
Hermann Ney. 1999.
\newblock Speech translation: Coupling of recognition and translation.
\newblock In \emph{Proceedings of ICASSP}, pages 517--520.

\bibitem[{Niehues and Cho(2017)}]{niehues2017exploiting}
Jan Niehues and Eunah Cho. 2017.
\newblock Exploiting linguistic resources for neural machine translation using multi-task learning.
\newblock In \emph{Proceedings of the Second Conference on Machine Translation}, pages 80--89.

\bibitem[{Post et~al.(2019)Post, Ding, Martindale, and Wu}]{post2019exploration}
Matt Post, Shuoyang Ding, Marianna Martindale, and Winston Wu. 2019.
\newblock An exploration of placeholding in neural machine translation.
\newblock In \emph{Proceedings of Machine Translation Summit XVII: Research Track}, pages 182--192.

\bibitem[{Post et~al.(2013)Post, Kumar, Lopez, Karakos, Callison-Burch, and Khudanpur}]{Post2013ST}
Matt Post, Gaurav Kumar, Adam Lopez, Damianos Karakos, Chris Callison-Burch, and Sanjeev Khudanpur. 2013.
\newblock Improved speech-to-text translation with the fisher and callhome spanish-english speech translation corpus.
\newblock In \emph{Proceedings of the 10th international workshop on spoken language translation: papers}.

\bibitem[{Prabhavalkar et~al.(2023)Prabhavalkar, Hori, Sainath, Schl{\"u}ter, and Watanabe}]{prabhavalkar2023end}
Rohit Prabhavalkar, Takaaki Hori, Tara~N Sainath, Ralf Schl{\"u}ter, and Shinji Watanabe. 2023.
\newblock End-to-end speech recognition: A survey.
\newblock \emph{IEEE/ACM Transactions on Audio, Speech, and Language Processing}, 32:325--351.

\bibitem[{Pradhan et~al.(2013)Pradhan, Moschitti, Xue, Ng, Bj{\"o}rkelund, Uryupina, Zhang, and Zhong}]{pradhan2013towards}
Sameer Pradhan, Alessandro Moschitti, Nianwen Xue, Hwee~Tou Ng, Anders Bj{\"o}rkelund, Olga Uryupina, Yuchen Zhang, and Zhi Zhong. 2013.
\newblock Towards robust linguistic analysis using ontonotes.
\newblock In \emph{Proceedings of the Seventeenth Conference on Computational Natural Language Learning}, pages 143--152.

\bibitem[{Sennrich and Haddow(2016)}]{sennrich2016linguistic}
Rico Sennrich and Barry Haddow. 2016.
\newblock Linguistic input features improve neural machine translation.
\newblock In \emph{Proceedings of the First Conference on Machine Translation: Volume 1, Research Papers}, pages 83--91.

\bibitem[{Song et~al.(2019)Song, Zhang, Yu, Luo, Wang, and Zhang}]{song2019code}
Kai Song, Yue Zhang, Heng Yu, Weihua Luo, Kun Wang, and Min Zhang. 2019.
\newblock Code-switching for enhancing nmt with pre-specified translation.
\newblock In \emph{Proceedings of the 2019 Conference of the North American Chapter of the Association for Computational Linguistics: Human Language Technologies, Volume 1 (Long and Short Papers)}, pages 449--459.

\bibitem[{Sperber and Paulik(2020)}]{sperber2020speech}
Matthias Sperber and Matthias Paulik. 2020.
\newblock Speech translation and the end-to-end promise: Taking stock of where we are.
\newblock In \emph{Proceedings of the 58th Annual Meeting of the Association for Computational Linguistics}, pages 7409--7421.

\bibitem[{Team et~al.(2023)Team, Anil, Borgeaud, Alayrac, Yu, Soricut, Schalkwyk, Dai, Hauth, Millican et~al.}]{team2023gemini}
Gemini Team, Rohan Anil, Sebastian Borgeaud, Jean-Baptiste Alayrac, Jiahui Yu, Radu Soricut, Johan Schalkwyk, Andrew~M Dai, Anja Hauth, Katie Millican, et~al. 2023.
\newblock Gemini: a family of highly capable multimodal models.
\newblock \emph{arXiv preprint arXiv:2312.11805}.

\bibitem[{Ugawa et~al.(2018)Ugawa, Tamura, Ninomiya, Takamura, and Okumura}]{ugawa2018neural}
Arata Ugawa, Akihiro Tamura, Takashi Ninomiya, Hiroya Takamura, and Manabu Okumura. 2018.
\newblock Neural machine translation incorporating named entity.
\newblock In \emph{Proceedings of the 27th International Conference on Computational Linguistics}, pages 3240--3250.

\bibitem[{Vila et~al.(2018)Vila, Escolano, Fonollosa, and Costa-Jussa}]{vila2018end}
Laura~Cross Vila, Carlos Escolano, Jos{\'e}~AR Fonollosa, and Marta~R Costa-Jussa. 2018.
\newblock End-to-end speech translation with the transformer.
\newblock In \emph{Proceedings of Interspeech}, pages 60--63.

\bibitem[{Wang et~al.(2023)Wang, Sun, Xue, Wu, Zhou, Gaur, Liu, and Li}]{wang2022lamassu}
Peidong Wang, Eric Sun, Jian Xue, Yu~Wu, Long Zhou, Yashesh Gaur, Shujie Liu, and Jinyu Li. 2023.
\newblock Lamassu: Streaming language-agnostic multilingual speech recognition and translation using neural transducers.
\newblock In \emph{INTERSPEECH 2023}, pages 57--61.

\bibitem[{Wang et~al.(2024)Wang, Wu, Caseiro, Munkhdalai, Sim, Rondon, Pundak, Song, Prabhavalkar, Meng et~al.}]{wang2024contextual}
Weiran Wang, Zelin Wu, Diamantino Caseiro, Tsendsuren Munkhdalai, Khe~Chai Sim, Pat Rondon, Golan Pundak, Gan Song, Rohit Prabhavalkar, Zhong Meng, et~al. 2024.
\newblock Contextual biasing with the knuth-morris-pratt matching algorithm.
\newblock In \emph{Proc. Interspeech 2024}, pages 282--286.

\bibitem[{Wang et~al.(2017)Wang, Cheng, Jiang, Yang, Chen, Li, Shi, Wang, and Yang}]{wang2017sogou}
Yuguang Wang, Shanbo Cheng, Liyang Jiang, Jiajun Yang, Wei Chen, Muze Li, Lin Shi, Yanfeng Wang, and Hongtao Yang. 2017.
\newblock Sogou neural machine translation systems for wmt17.
\newblock In \emph{Proceedings of the Second Conference on Machine Translation}, pages 410--415.

\bibitem[{Xie et~al.(2022)Xie, Xia, Wu, Huang, Fan, and Qin}]{xie2022end}
Shufang Xie, Yingce Xia, Lijun Wu, Yiqing Huang, Yang Fan, and Tao Qin. 2022.
\newblock End-to-end entity-aware neural machine translation.
\newblock \emph{Machine Learning}, 111(3):1181--1203.

\bibitem[{Xue et~al.(2022)Xue, Wang, Li, Post, and Gaur}]{xue2022large}
Jian Xue, Peidong Wang, Jinyu Li, Matt Post, and Yashesh Gaur. 2022.
\newblock Large-scale streaming end-to-end speech translation with neural transducers.
\newblock \emph{arXiv preprint arXiv:2204.05352}.

\bibitem[{Xue et~al.(2023)Xue, Wang, Li, and Sun}]{xue2023weakly}
Jian Xue, Peidong Wang, Jinyu Li, and Eric Sun. 2023.
\newblock A weakly-supervised streaming multilingual speech model with truly zero-shot capability.
\newblock In \emph{2023 IEEE Automatic Speech Recognition and Understanding Workshop (ASRU)}, pages 1--7. IEEE.

\bibitem[{Yang et~al.(2023)Yang, Gu, Liu, Ghosh, Bulyko, and Stolcke}]{yang2023generative}
Chao-Han~Huck Yang, Yile Gu, Yi-Chieh Liu, Shalini Ghosh, Ivan Bulyko, and Andreas Stolcke. 2023.
\newblock Generative speech recognition error correction with large language models and task-activating prompting.
\newblock In \emph{2023 IEEE Automatic Speech Recognition and Understanding Workshop (ASRU)}, pages 1--8. IEEE.

\bibitem[{Zhao et~al.(2019)Zhao, Sainath, Rybach, Rondon, Bhatia, Li, and Pang}]{zhao2019shallow}
Ding Zhao, Tara~N Sainath, David Rybach, Pat Rondon, Deepti Bhatia, Bo~Li, and Ruoming Pang. 2019.
\newblock Shallow-fusion end-to-end contextual biasing.
\newblock In \emph{Interspeech}, pages 1418--1422.

\bibitem[{Zhao et~al.(2024{\natexlab{a}})Zhao, Zhang, Yu, Wang, Geng, Fu, Yang, Zhang, Jiang, and Cui}]{zhao2024retrieval}
Penghao Zhao, Hailin Zhang, Qinhan Yu, Zhengren Wang, Yunteng Geng, Fangcheng Fu, Ling Yang, Wentao Zhang, Jie Jiang, and Bin Cui. 2024{\natexlab{a}}.
\newblock Retrieval-augmented generation for ai-generated content: A survey.
\newblock \emph{arXiv preprint arXiv:2402.19473}.

\bibitem[{Zhao et~al.(2024{\natexlab{b}})Zhao, Li, Fan, and Post}]{zhao2024ctc}
Rui Zhao, Jinyu Li, Ruchao Fan, and Matt Post. 2024{\natexlab{b}}.
\newblock Ctc-gmm: Ctc guided modality matching for fast and accurate streaming speech translation.
\newblock In \emph{2024 IEEE Spoken Language Technology Workshop (SLT)}, pages 1068--1075. IEEE.

\end{thebibliography}

\newpage

\end{document}